\documentclass[runningheads]{llncs}
\usepackage{graphicx}
\usepackage{comment}
\usepackage{amsmath,amssymb} 
\usepackage{color}

\usepackage[width=122mm,left=12mm,paperwidth=146mm,height=193mm,top=12mm,paperheight=217mm]{geometry}

\usepackage[T1]{fontenc}    
\usepackage[utf8]{inputenc} 
\usepackage[pagebackref=true,breaklinks=true,colorlinks,bookmarks=false]{hyperref}
\usepackage{url}            
\usepackage{amsfonts}       
\usepackage{tabulary,multirow,xcolor}
\usepackage{microtype}

\usepackage{tabularx}
\usepackage{booktabs}       
\newcommand{\citet}[1]{\cite{#1}}

\newcommand{\ve}[1]{\mathbf{#1}} 

\newcommand{\done}[1]{}

\newcommand{\app}{\raise.17ex\hbox{$\scriptstyle\sim$}}

\def\x{$\times$}
\newcolumntype{x}[1]{>{\centering\arraybackslash}p{#1pt}}

\newlength\savewidth\newcommand\shline{\noalign{\global\savewidth\arrayrulewidth
  \global\arrayrulewidth 1pt}\hline\noalign{\global\arrayrulewidth\savewidth}}
\newcommand{\tablestyle}[2]{\setlength{\tabcolsep}{#1}\renewcommand{\arraystretch}{#2}\centering\footnotesize}

\makeatletter\renewcommand\paragraph{\@startsection{paragraph}{4}{\z@}
  {.5em \@plus1ex \@minus.2ex}{-.5em}{\normalfont\normalsize\bfseries}}\makeatother

\begin{document}
\pagestyle{headings}
\mainmatter

\title{Beyond the Camera: \\Neural Networks in World Coordinates}

\titlerunning{Beyond the Camera: Neural Networks in World Coordinates} 
\authorrunning{G. A. Sigurdsson et al.}
\newcommand{\myurl}[1]{\url{#1}}

\renewcommand{\thefootnote}{\fnsymbol{footnote}}
\author{Gunnar A. Sigurdsson\inst{~1\star}  
\\ Abhinav Gupta\inst{~1} ~~~
Cordelia Schmid\inst{~2} ~~~ 
Karteek Alahari\inst{~2} \\
\vspace{0.5em}
\inst{1}Carnegie Mellon University~~~~ \  
\inst{2}Inria\inst{\star\star}\\
\myurl{github.com/gsig/world-features}\vspace{-0.5cm}
}
\institute{}

\maketitle

\begin{abstract}
Eye movement and strategic placement of the visual field onto the retina, gives animals increased resolution of the scene and suppresses distracting information. This fundamental system has been missing from video understanding with deep networks, typically limited to 224 by 224 pixel content locked to the camera frame. We propose a simple idea, WorldFeatures, where each feature at every layer has a spatial transformation, and the feature map is only transformed as needed. We show that a network built with these WorldFeatures, can be used to model eye movements, such as saccades, fixation, and smooth pursuit, even in a batch setting on pre-recorded video. That is, the network can for example use all 224 by 224 pixels to look at a small detail one moment, and the whole scene the next. We show that typical building blocks, such as convolutions and pooling, can be adapted to support WorldFeatures using available tools.
Experiments are presented on the Charades, Olympic Sports, and Caltech-UCSD Birds-200-2011 datasets, exploring action recognition, fine-grained recognition, and video stabilization.
\end{abstract}

\footnotetext[1]{Work was done while Gunnar was at Inria.}
\footnotetext[2]{Univ.\ Grenoble Alpes, Inria, CNRS, Grenoble INP, LJK, 38000 Grenoble, France.}

\section{Introduction}

The success of recent vision systems is in a way surprising, since the input is typically a $224 {\times} 224$ pixel image, or in the case of videos, a temporal stack of such images~\cite{i3d,feichtenhofer2018slowfast}. This is both low resolution, and does not give much flexibility to investigate important signals in the image. The system is constrained to what the camera recorded, and has typically no active role in collecting the data. In particular, video recognition architectures have been shown to be vulnerable to camera motion, subject size, and temporal scale~\cite{sigurdsson2017actions}. In comparison, the human visual system is not a passive receiver---our eyes are constantly moving to increase the effective resolution of the scene and suppress irrelevant signals~\cite{findlay2009saccadic}.

\begin{figure}[t]
    \centering
    \includegraphics[width=1.0\linewidth]{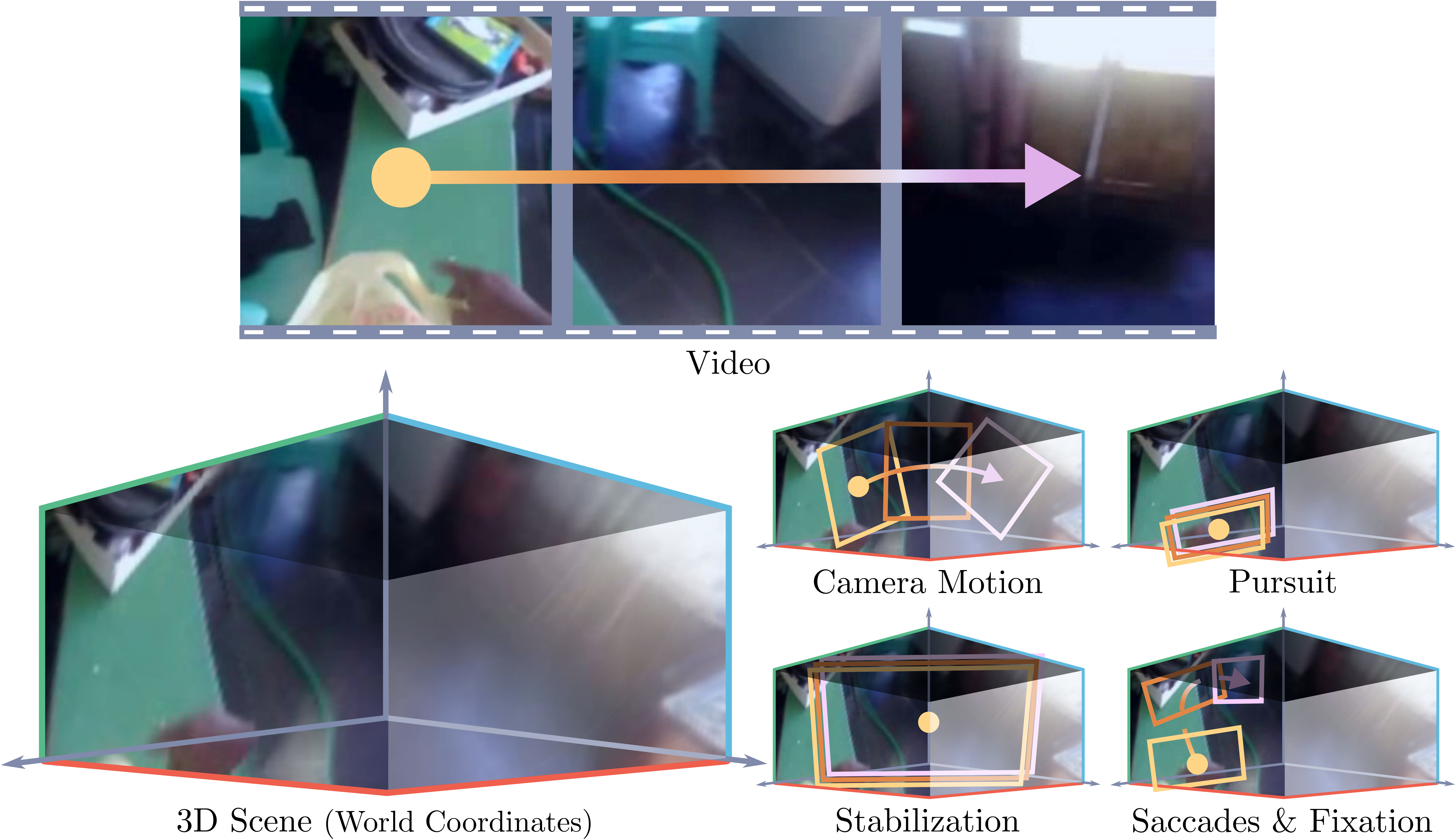}
    \caption{In this paper we explore how to model the variety of eye motions, even on pre-recorded data. We generalize the concept of feature maps, to WorldFeatures, that include a relative location in space and time, which can encode any type of ``eye motion'' we want apply to the recorded data, or undo. Orange to pink is used to denote features at time $t{-}1$, $t$, and $t{+}1$. 
    We show a video from a first-person view, along with the \emph{Camera Motion} in the reconstructed world frame. The original video follows a particular sequence of viewpoints, however, crucial information about the activities in the video requires different views---\emph{Stabilization} transforms the views to the world frame, \emph{Pursuit} transforms the views to follow the subject of the video, and \emph{Saccades \& Fixation} scans the scene to extract fine-grained information from the video.
    }
    \label{fig:teaser}
\end{figure}

These eye movements exist in humans and animals with foveal vision, and can be categorized as: \emph{Stabilization}, the vestibulo-ocular reflex describes a control system the human visual system uses to stabilize images on the retina given proprioceptive information, like head rotation~\cite{crawford1991axes}. \emph{Smooth Pursuit}, where the gaze is voluntarily shifted to track a moving object, effectively stabilizing the object of interest on the retina~\cite{krauzlis1994temporal}. \emph{Fixation}, where gaze is fixed towards a single location, to enhance resolution of that area~\cite{rucci2015control}. \emph{Saccades}, where the eyes quickly jerk between phases of fixation to explore the scene~\cite{saccades}. In Fig.~\ref{fig:teaser} we illustrate how a video is constrained to what the camera recorded, but that also the video may be refocused. How can we build a vision system that has this flexibility to freely explore the data, and move beyond a fixed $224 {\times} 224$ window?

\begin{figure}[tb]
    \centering
    \includegraphics[width=0.8\linewidth]{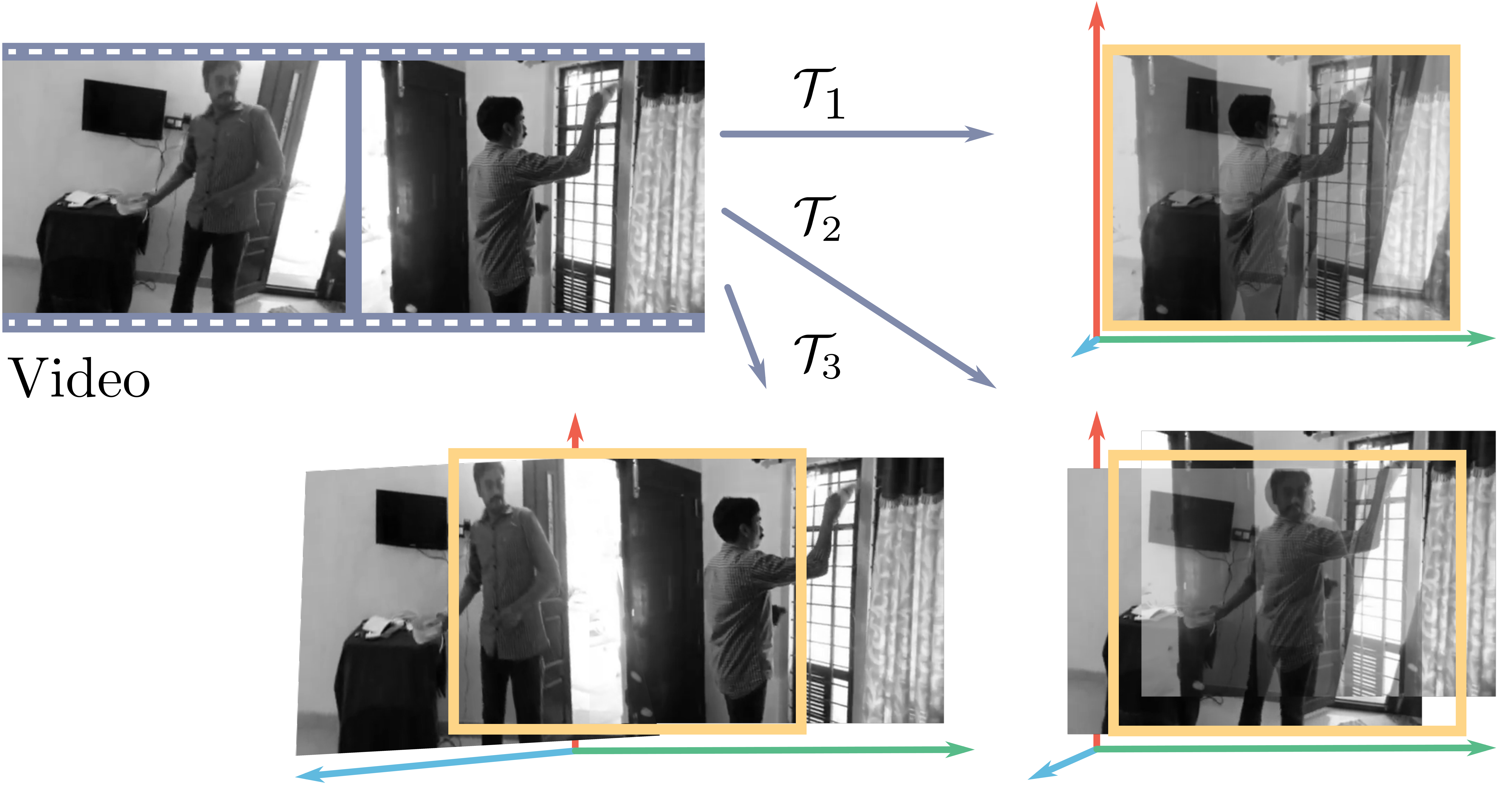}
    \caption{Illustration of WorldFeatures, attaching transformations to features. Top left is the original video shown as two frame featuremap. We show the temporal average of an aligned feature map, with three different transformations $\mathcal{T}_1$ (identity), $\mathcal{T}_2$ (pursuit), $\mathcal{T}_3$ (stabilization). Observe how different transforms highlight different aspects of the data. We show a potential coordinate frame, that could be used to fit the data into a network, highlighting a problem with naively stabilizing the data, for example, when 64 frames in a row need to be aligned. WorldFeatures get around this by keeping the transformation and only using it as needed.}
    \label{fig:worldfeature}
\end{figure}

We propose a simple and effective idea---each feature, has a location in real-world coordinates. This combination of \texttt{(feature, transformation)} pairs is referred to as a WorldFeature. An example transformation might be image coordinates to real-world coordinates (camera matrix). It turns out preprocessing the video data to, for example, stabilize it, introduces new problems, as demonstrated in Fig.~\ref{fig:worldfeature}. Instead, we use these WorldFeatures in all layers in the network and index the data according to the transformation or transform it as needed. That is, \emph{implicit} transformation instead of \emph{explicit}.
With such network, we can utilize any type of ``eye motion'' we want apply to the recorded data, or undo. 

To illustrate the idea, in Fig.~\ref{fig:worldfeature} we present a simple two frame featuremap with three different transformations $\mathcal{T}_1$, $\mathcal{T}_2$, $\mathcal{T}_3$ to showcase how different transformations can highlight different signals in the data. For example, in the third it is easy to understand the global layout of the scene, and in the second it is easier to understand the difference in the pose of the person. In Sec.~\ref{sec:WorldConvolutions} we show more details on WorldFeatures, and how to keep track of the transformations after every layer to build networks with these operations.

\paragraph{Contributions.}
In this paper we propose the idea of WorldFeatures, where each feature has a transformation, on multiple vision tasks: Fine-grained Image Recognition (Sec.~\ref{sec:finegrained}), Video Stabilization (Sec.~\ref{sec:stabilization}), and Video Activity Recognition (Sec.~\ref{sec:activityrecognition}). 
We formalize WorldFeatures, and show how different eye movements can be modeled in this framework. Further, we propose a simple implementation that can adapt optimized neural networks tools to operate on WorldFeatures. This implementation can extend many state-of-the-art building blocks that operate over space and time, such as 3D Convolutions, and MaxPooling. The output of each building block, is also WorldFeatures, such that it can be processed hierarchically (e.g. 3D-ResNet~\cite{wang2018non}).

\section{Related Work}
\label{sec:related}

Many image and video understanding architectures have been proposed in the last few decades and we refer the reader to~\cite{poppe2010survey,weinland2011survey} for a detailed survey, and focus here on recent video architectures related to our approach. 
\paragraph{Video Architectures.}
Recently, networks have become more evolved for separating content and motion~\cite{i3d,ryoo2013first,simonyan2014twostream,feichtenhofer2018slowfast}. This direction is analogous to the different specialization of the ventral and dorsal streams in the mammalian brain. Similarly, separating camera motion and subject motion has shown great promise in video segmentation~\cite{vijayanarasimhan2017sfm}. As the field moves towards utilizing longer temporal context~\cite{DBLP:journals/corr/abs-1812-05038}, equipping these networks with the capability to separate camera motion, object permanence, scene dynamics, and subject motion can greatly improve those efforts. This paper aims to provide the building blocks necessary to construct these systems.

\paragraph{Video Stabilization.}
Related to our work on understanding eye motion, is the task of video stabilization that has been pursued directly~\cite{grundmann2011auto,wang2018deep,liu2014steadyflow}, or with additional supervision~\cite{liu2012video}. Of particular relevance is recent work that has incorporated stabilization and tracking of points over time to improve accuracy of video recognition~\cite{wang2011action,wang2015action}. Inspired by these successes, we go a step further, and build a framework that can operate given an arbitrary transformation signal, and use this to improve the vision system.

\paragraph{Improving Visual Input.}
On the other hand, recent work has also explored how to use the input data more effectively. This has been done through spatially or temporally modifiable connections~\cite{dai2017deformable}, non-local connections where feature similarity guides the connections~\cite{wang2018non}, or transformations of the input~\cite{jaderberg2015spatial}. 
Furthermore,~\citet{recasens2018learning} investigated the idea that not all input features are equally useful, and explored how to learn how to highlight and ``zoom in'' on the important content in each image. As we will see, eye movement is primarily useful to allow for efficient allocation of resources, and since our system can handle arbitrary transformations, we can explicitly utilize various ``zoom in'' of the data. Furthermore, our WorldFeatures is a system that allows a network to generalize beyond the camera, and is complementary to frameworks that learn attention to emphasize parts of the data.

\section{WorldFeatures: Feature Maps with Camera Movements}
\label{sec:WorldConvolutions}

We start by introducing the ideas behind WorldFeatures, that is, \texttt{(feature, transformation)} pairs. Then we demonstrate how different eye movements can be modeled in this framework. A video is a specific window into the world at the time it was recorded, and a system analyzing this video might have a completely different objective for watching the scene than the camera operator. To undo the camera bias or highlight signals in the video, we allow a transformations that are analogous to eye movements, after the data has already been recorded.

To undo camera transformations, could we pre-process the data? For example, applying affine transformations to each frame to undo camera movement, for stabilization. In fact, even this simple case has problems: How to fit this transformed data into the model? At what scale should we process the data? How to use imperfect transforms? Consider eyes scanning a scene, fixating on a variety of points in the scene at multiple resolutions. Creating data that follows a similar pattern without an understanding how they relate to each other creates a noisy and disjointed view of the scene. Instead, by keeping track of the relative transformations between features, we can crop, scale, and zoom in on the data as required, while still maintaining correct relationships between features.
\begin{figure}[tb]
    \centering
    \includegraphics[width=1.0\linewidth]{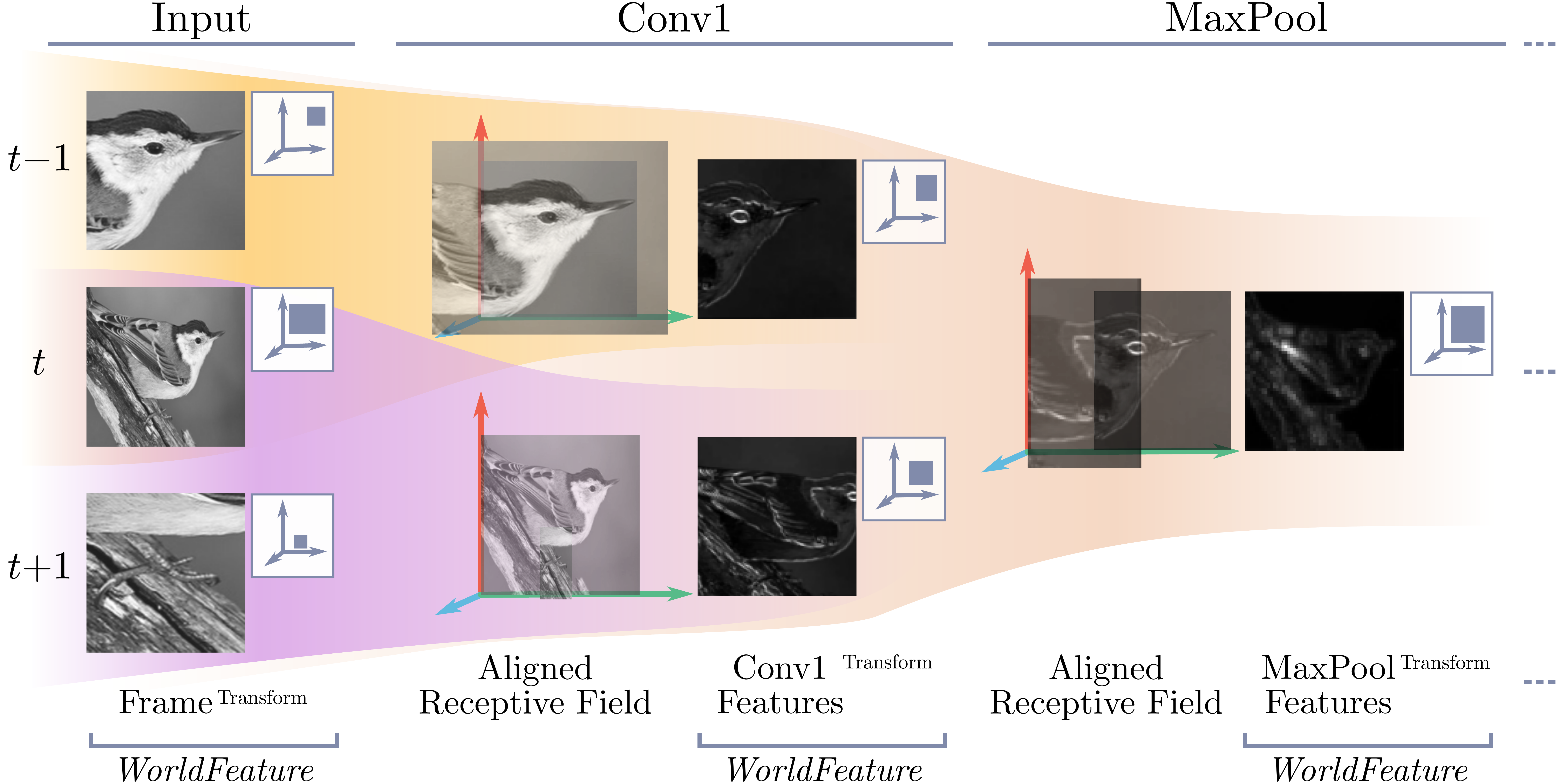}
    \caption{Demonstration of a model using WorldFeatures at each layer. Here we look at an illustration of the outputs of conv1 and maxpool layers, when applied to an ``eye movement'' task. The input and output of each layer are WorldFeatures. The receptive field of each output featuremap is aligned, explicitly or implicitly depending on implementation, before computing the output. Since each output feature map has a different alignment, each output has a different transformation, and we repeat the process at every layer. This shows that WorldFeatures allow the network to combine information across multiple locations and scales in the image.}
    \label{fig:convstack}
\end{figure}

\subsection{WorldFeatures Definition}
The key to our idea is the concept of WorldFeatures $\ve{x}^\mathcal{T}$. That is, a feature map $\ve{x}$ of size $T {\times} H {\times} W$ (time, height, width) and an arbitrary transform $\mathcal{T}$ which can encode any type of ``eye motion'' we want to apply to the recorded data---or undo.\footnote{For example, the affine transform, $\mathcal{T}(t{,}h{,}w) = (t,\ a^0_th{+}a^1_tw{+}a^2_t,\ a^3_th{+}a^4_tw{+}a^5_t)$, where $a^n_t$ are the parameters at frame $t$.}
We refer to a single $H {\times} W$ timepoint as a frame, and note that $\ve{x}{=}\ve{x}^\mathcal{I}$, where $\mathcal{I}$ is the identity transform.
The feature of $\ve{x}^\mathcal{T}$ at $(t{,}h{,}w)$ is written $\ve{x}\left(\mathcal{T}(t{,}h{,}w)\right)$. Therefore, extending a convolution operation to work on $\ve{x}^\mathcal{T}$:
\begin{align}
\ve{y}^\mathcal{T'}(t,h,w) &= \sum_i \sum_j \sum_k \ve{x}(\mathcal{T}(i,j,k)) \cdot \ve{f}(t{-}i,h{-}j,w{-}k)\,,\label{eq:sconv}
\end{align}
where $\ve{y}^\mathcal{T'}$ is also a WorldFeature, that can then be passed to the next convolution operation and so on. Here $\ve{f}$ is a filter, and we omit batch and channel dimensions for clarity. 
Compared to regular convolutions, the transformation $\mathcal{T}$ is necessary to transform the features as needed.
For example, the valid values of each feature map are only defined for $t {\in} \left[0,T\right)$, $h {\in} \left[0,H\right)$, and $w {\in} \left[0,W\right)$, and transforming the video before processing would move content out of the frame, etc. 
This convolution layer could use a Spatial Transformer Network~\cite{jaderberg2015spatial} independently transforming the receptive field for each output frame, at every layer, exponentially increasing computation. This implementation of WorldFeatures is illustrated in Fig.~\ref{fig:convstack}. Fortunately, in Sec.~\ref{sec:experiments} we outline an efficient implementation of this for CNN building blocks, which allows us to explore this type of networks.

\subsection{Transformation Types}
This framework can be used to model eye movements: Stabilization, Smooth Pursuit, Fixation and Saccades. Fig.~\ref{fig:teaser} contains illustrative examples for each that we refer to for each of these transformations.

\paragraph{Stabilization.}
The first and most intuitive transformation $\mathcal{T}$ we encounter is the stabilizing transform. We obtain the camera matrix for every frame, and invert it to obtain the stabilizing transform. That is, the transformed feature map $\ve{x}\left(\mathcal{T}(t{,}h{,}w)\right)$, will contain the same point in the world at location $(h{,}w)$ for any $t$. The \emph{Stabilization} example in Fig.~\ref{fig:teaser} visualizes how the frames of the given video are arranged in the world frame.

\paragraph{Smooth Pursuit.}
Since the transform $\mathcal{T}$ is arbitrary we can also choose it to stabilize with respect to a different point of reference. That is, instead of $(h{,}w)$ for any $t$ pointing to the same point in the world, we can make it point to the same point on a moving object of interest, such as a person. The \emph{Pursuit} example in Fig.~\ref{fig:teaser} visualizes how the frames of the given video can be arranged, and combined with attention to the hands of the person.

\paragraph{Fixation.}
To model fixation, where the high-resolution part of the retina (the fovea) is used, we use a transform that enhances part of the frame. That is, if $\ve{x}^\mathcal{T}(t{,}h{,}w)$ is the original WorldFeature, then we add a ``fixation transform'' $\mathcal{F}$ to get the WorldFeature $\ve{f}^{\mathcal{T} \circ \mathcal{F}}$ where $\ve{f} = \ve{x}\left(\mathcal{F}(t{,}h{,}w)\right)$. This fixation transform ``zooms in'' on a part of the input, increasing the resolution.

\paragraph{Saccades / Visual Search.}
To model saccades, we proceed similarly to fixation, and define a transform $\mathcal{S}$ that gazes at a particular part of the frame, allowing the model to attend to different parts of the image. The similarity with the implementation of fixation is because saccades are in fact defined as the jump between fixations. In the \emph{Saccades \& Fixation} example in Fig.~\ref{fig:teaser} we demonstrate how the transformations can be chosen to pay close attention to important aspects of the scene, such as the hands, objects and context.

\paragraph{}
In conclusion, there are multiple eye movements possible, allowing a variety of augmentations without sacrificing relative location between features. In the experiments we use these definitions to provide the model with a transformation for each video, but learned transformations could be used as well.

\section{Experiments.}
\label{sec:experiments}

Here we explore WorldFeatures for diverse applications. We use a 3DResNet-style architecture for videos built from WorldFeatures. This can be used to incorporate arbitrary eye movements in videos and images, where images are just treated as an uneventful video (replicating images in time). The network can then utilize the transformations to ``zoom in'' on parts of the image. First, we demonstrate that a 3D network, can outperform 2D networks on fine-grained recognition of nature categories (image task). Second, we look at video stabilization, and show how implicit transformations can bypass the problems with explicitly stabilizing the input data with preprocessing. Third, we utilize enhancing transformations, smooth pursuit and fixation, to improve video recognition accuracy.

\begin{figure}[tb]
    \centering
    \includegraphics[width=0.8\linewidth]{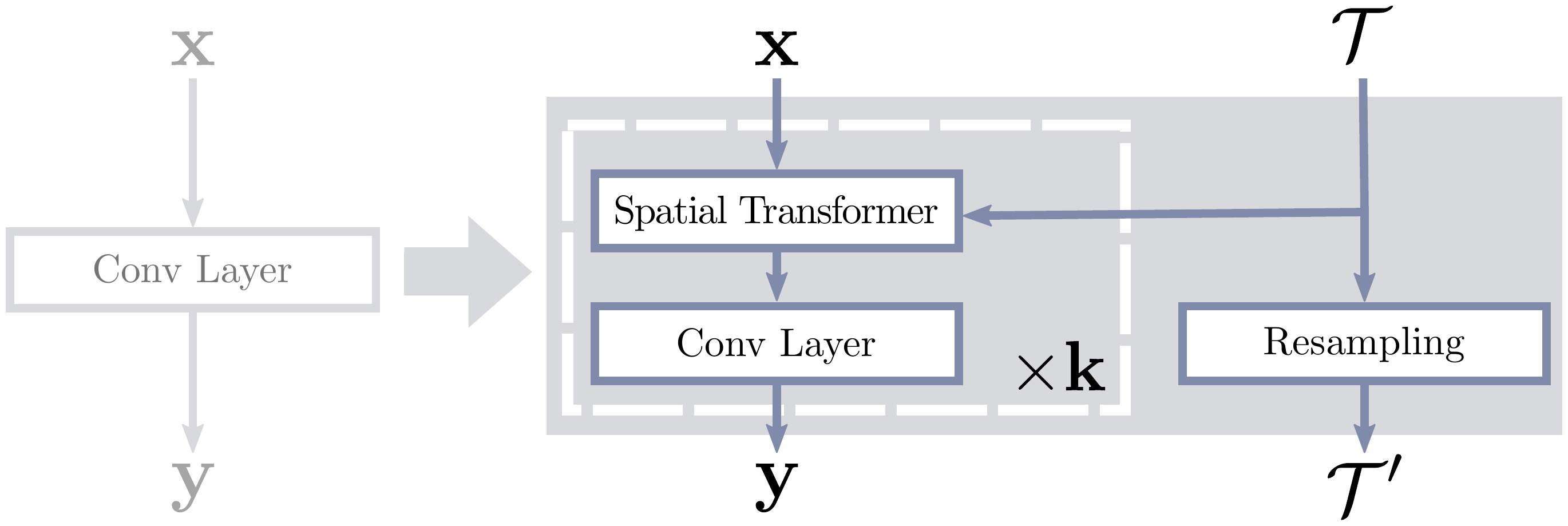}
    \caption{Our simple implementation of WorldFeatures ``wraps'' a layer (e.g. conv layer) and applies a spatial transformer to the data to align the receptive field of each output, $k$ times for a layer with temporal extent of $k$ (e.g. 3 in ResNet), and combines the outputs.\protect\footnotemark[4]{} More details are provided in the Appendix.}
    \label{fig:modelfig}
\end{figure}

\paragraph{Implementation.}
Our implementation extends a 3DResNet50 model, following~\citet{wang2018non} (similar to I3D~\cite{i3d}). All convolution, max pool, and average pool layers are converted to WorldFeature layers, which implicitly operate on a transformed video. All models start with parameters from a regular 3DResNet50 pretrained on the Kinetics dataset~\cite{i3d}. 
In Fig.~\ref{fig:modelfig} we illustrate a simple method to convert a layer, such as a conv layer, to use WorldFeatures, without having to engineer special low-level GPU programs for each. More details are provided in the Appendix.
We use the grid sampler from Spatial Transformer Networks~\cite{jaderberg2015spatial}, with arbitrary transformation grids. After fitting the transformer, we use the grid sampler in the nearest (pixel) setting. This avoids blurring of the feature maps as they are repeatedly transformed. To address missing values after transforming, we add a channel denoting missing data at a point, similar to~\citet{felzenszwalb2009object}.
For comparison, all methods use the same batch-size of 2 unless otherwise noted, and follow the same learning rate schedule tuned for the baseline. The models are implemented using the PyVideoResearch~\cite{sigurdsson2018pyvideoresearch} framework in PyTorch, and will be available at \myurl{github.com/gsig/world-features}.
\footnotetext[4]{For $k{=}3$ the first transformer aligns frames $\{0{,} 1{,} 2\}$ (the receptive field for output frame $1$), and frames $\{3{,} 4{,} 5\}$ (receptive field for output frame $4$), etc. The second transformer aligns $\{1{,} 2{,} 3\}$ (receptive field for output frame $2$), etc.}

\subsection{Saccades - Fine-Grained Recognition}
\label{sec:finegrained}
\begin{figure}[tb]
    \centering
    \hspace{-1.45em}
    \includegraphics[width=1.03\linewidth]{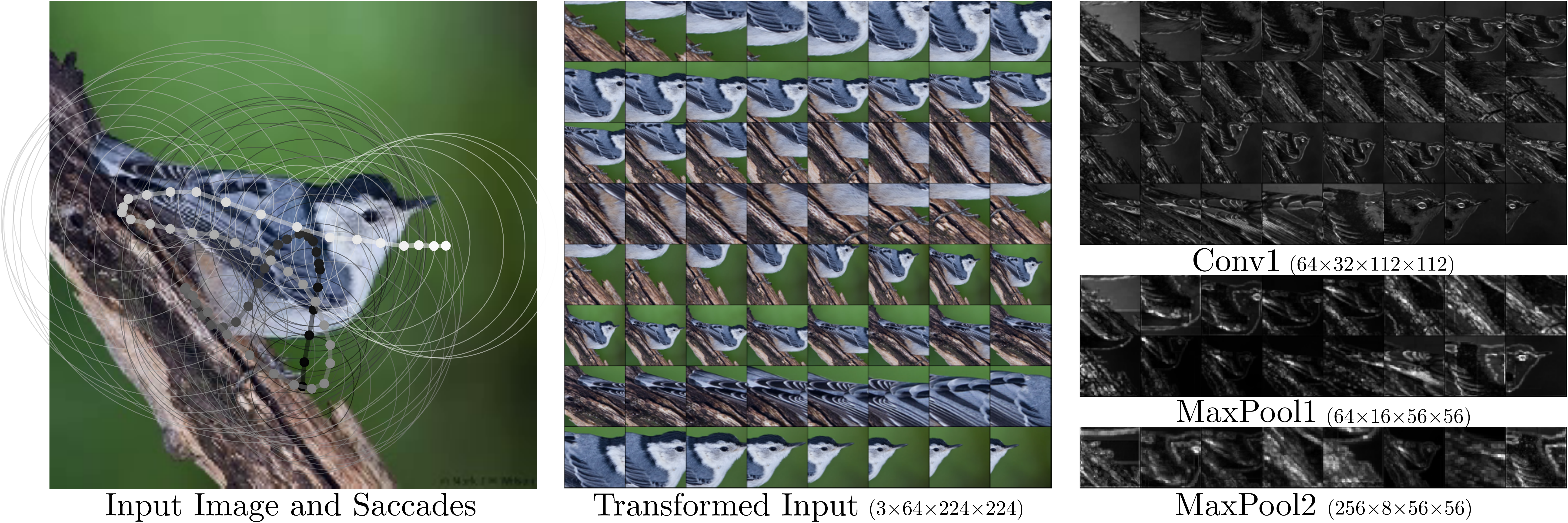}
    \caption{In the left column we visualize the path (the saccades) that the model chooses, given static saliency and objectness, where dark to light denotes frames 1 through 64, and the large circles denote the scale of the fixaton in that frame. In the the middle column, we show the pixel data that the model uses (the model also receives the transformations). On the right, we show featuremaps from Conv1, MaxPool1, and MaxPool2, high values denote high variance over time at that pixel.}
    \label{fig:bird}
\end{figure}

Motivated by human gaze studies in image recognition, we explore using 3D video networks for fine-grained recognition on the Caltech-UCSD Birds-200-2011 dataset~\cite{WahCUB_200_2011}. To provide a gaze trajectory for the network, we use saliency~\cite{hou2007saliency} and objectness~\cite{cheng2014bing} to generate 64 bounding boxes. We replicate each 2D image 64 times in time to form a 3D input. The 64 bounding boxes form the basis for the fixation in each frame, and are ordered to maximize the overlap between consecutive boxes, allowing the network to learn filters that combine information between various fixations and scales. See Fig.~\ref{fig:bird} for an example. We compare with regular \emph{3DResNet50}, and \emph{ResNet50}. We start with a 3DResNet50 and fine-tune with the saccades/fixation setup, and combine with ResNet50.
\begin{table}[tbb]
\centering
\begin{tabularx}{\linewidth}{@{\hskip 1em}r@{\hskip 1cm}lX@{\hskip 1em}c@{\hskip 1em}}
\toprule & Top-1 (\%) && Resolution (px.)\\ \midrule
3DResNet50 & 79.8 && 224 \\
DT-RAM~\cite{li2017dynamic} & 82.8 && 227 \\
ResNet50 & 83.4 && 224 \\
Learning to Zoom~\cite{recasens2018learning} & 84.5 && 227 \\
3DResNet50 w/Saccades & \textbf{85.3} && 224 \\ \midrule
DT-RAM~\cite{li2017dynamic} & 86.0 && 448 \\ \bottomrule
\end{tabularx}
\caption{Fine-grained Recognition on the Caltech-UCSD Birds-200-2011 dataset.}
\label{tab:saccades}
\end{table}

\paragraph{Results.}
The results are presented in Table~\ref{tab:saccades}. With a input size of 224 pixels, our method outperforms architectures specialized for fine-grained recognition, even with recent methods such as~\citet{recasens2018learning} that learn the function of what to zoom to, whereas we use off-the-shelf saliency. Combining our framework and learning saliency is likely to yield additional gains. The method from~\cite{li2017dynamic} only outperforms other methods when provided with $448{\times}448$ image requiring a special network beyond the scope of this work. 
These results demonstrate an exciting new avenue for video architectures applied to the conventionally 2D task of fine-grained recognition.
In Fig.~\ref{fig:bird} we visualize the points of interests and scales suggested by the saliency algorithm, and how that is translated into pixel data that the network sees (along with the transformations between those). We also show example of featuremaps from 3 layers in the network, where we can see how the network combines information across the different saccades, and builds detailed fine-grained understanding of the image.

\subsection{Stabilization - Video Activity Recognition}
\label{sec:stabilization}

Stabilization is a fundamental vision task, and has been explored in various contexts. Here we specifically explore stabilization for improving video recognition.

To evaluate the stabilization performance, we start with videos with synthetic camera motion. We use $50\%$ of the Charades~\cite{charades} videos with the least motion\footnote{We used variance of optical flow, yielding 4396 videos of average 30 sec.}, and add synthetic camera movement by linearly moving between 3 locations at randomly chosen scale levels ($30\%$--$300\%$, \emph{Synthetic}). Next we consider Olympic-Sports~\cite{niebles2010modeling}, that was used to showcase the advantage of stabilization in the seminal IDT work~\cite{wang2011action}. 
Charades is evaluated with a video-level mean average precision over 157 classes, and Olympic-Sports uses a top1 accuracy for each video over 20 classes.
Our baselines are \emph{3DResNet50}~\cite{wang2018non},
and 3DResNet50 with all frames are stabilized to the center frame (\emph{3DResNet50++}). All models use a 16 temporal frame stack as input to make stabilization feasible for the baseline, and are trained with the same learning rate schedule and hyperparameters tuned to the baseline. For each video we construct the camera motion from the video using direct optimization of affine transformations parameters between consecutive frames. The details are provided in the Appendix.

\paragraph{Results.}
In Table~\ref{tab:stabilization} we see that using the stabilization transformation allows our method to improve over the baselines. 
That is, modelling features and transformations together instead of preprocessing allows our method to avoid compounding errors do to stabilization over many frames, and loss of resolution.
\begin{table}[tbh]
\centering
\caption{Results on stabilization on two video activity recognition datasets. Stabilizing the video before inputting it to the network (\emph{3DResNet50++}) helps in some cases, but has drawbacks, whereas \emph{implicit} stabilization can better utilize the input.}
\label{tab:stabilization}
\begin{tabularx}{\linewidth}{@{\hskip 1em}Xc@{\hskip 2em}c@{\hskip 2em}c@{\hskip 1em}}
\toprule & 3DResNet50 & 3DResNet50++ & Pursuit \\ \midrule
Synthetic (Video mAP) & 13.1 & 13.5 & \textbf{14.1} \\
Olympic-Sports (\% Top-1) & 96.4 & 97.2 & \textbf{97.2} \\
\bottomrule
\end{tabularx}
\end{table}

\paragraph{Analysis.}
We also looked at an egocentric video dataset, Charades-Ego~\cite{sigurdsson2018actor}.
With 64 frame inputs, preprocessing (\emph{3DResNet50++}) actually reduces performance in the baseline ($23.6\rightarrow22.8$ mAP), and with 16 frame inputs (only 0.67 sec video clip), preprocessing only slightly improves performance ($17.5\rightarrow18.5$ mAP). This demonstrates the challenge of stabilizing a video before passing it to the network, and interestingly, stabilization does not seem to help activity recognition much. Perhaps, since what the camera operator is looking at is highly informative in first-person video. In any case, WorldFeatures have performance at par with the better method in each case ($23.4$ and $18.4$ mAP).

\begin{figure}[tb]
    \centering
    \includegraphics[width=1.0\linewidth]{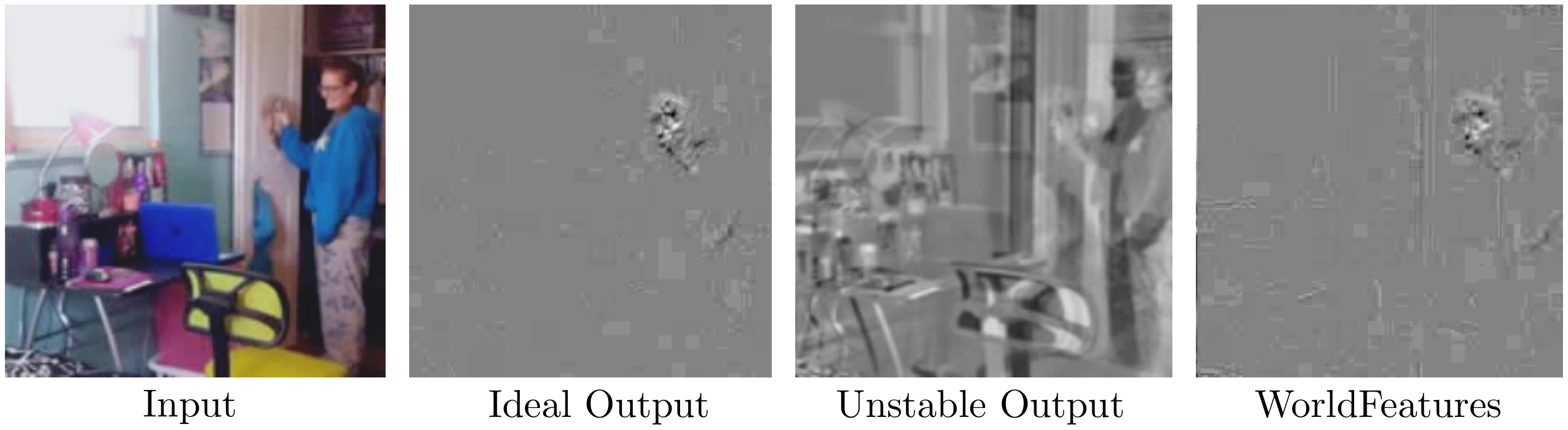}
    \caption{A single frame from 4 feature maps. First feature map is original video, second is the output of a temporal difference filter applied to an ideal (stable) video, third is output of the filter applied to the actual unstable video, and finally the output of our method given the estimated stabilizing transform.}
    \label{fig:temporaldifference}
\end{figure}

In Fig.~\ref{fig:temporaldifference} we demonstrate how camera motion makes learning from video challenging. We show point in time from 4 feature maps. We observe that WorldFeatures can utilize a transformation to suppress irrelevant signals. Although a deep network might, with sufficient data, be able to learn all possible variations, this adds complexity the model. We will see in the next section how we can go even further and use the transformations to focus on particular parts of the data.

\subsection{Smooth Pursuit \& Fixation - Video Activity Recognition}
\label{sec:activityrecognition}

To demonstrate how WorldFeatures can utilize any transformations, we explored \emph{Smooth Pursuit}, where we use a transformation to stabilize the content with respect to a human bounding box (\emph{Pursuit}). We use an RCNN person detector~\cite{renNeurIPS15fasterrcnn}, and construct a the transform such that the person is always centered and fills the input to the network. 
Going further, we explored how to utilize the transformations to increase the effective resolution of particular parts of the scene or the subject in a video. Concretely, we used simple video saliency (temporal difference), and fit a bounding box around $80\%$ of the variance in the saliency, and the fixation transforms the input such that the box fills the input (\emph{Fixation}). 

Our experiments are on the Charades dataset~\cite{charades}, evaluated with a video-level mean average precision over 157 classes. We start with a standard pre-trained baseline network from the literature (\emph{3DResNet50}), fine-tune with the new framework, and combine with the old network.

\paragraph{Results.}
Our results are presented in Table~\ref{tab:fixation}. Interestingly, adding smooth pursuit on its own does not significantly improve over the baseline, which stabilizes the video with respect to the person. However, this is expected since many of the videos in Charades have camera motion that already follows the video subject. Fortunately, we do see that using smooth pursuit followed by fixation (\emph{FixationPursuit}) we can improve performance. This suggests that whereas stabilization on its own is not particularly helpful to current neural networks, when it is used to locate an area to increase the resolution it can offer significant benefits, since $224{\times}224$ pixels is not provide much detail, and learn filters that combine information across multiple locations and resolutions.
\begin{table}[tbh]
\centering
\caption{Activity Recognition on the Charades dataset using smooth pursuit and fixation transformations. \label{tab:fixation}}
\begin{tabularx}{\linewidth}{@{\hskip 1em}Xc@{\hskip 1em}c@{\hskip 1em}c@{\hskip 1em}c@{\hskip 1em}}
\toprule & 3DResNet50 & Pursuit & Fixation & FixationPursuit \\ \midrule
Charades~\cite{charades} (Video mAP) & 31.3 & 31.5 & 32.3 & \textbf{32.6} \\
\bottomrule
\end{tabularx}
\end{table}

\begin{figure}[tb]
\centering
\includegraphics[width=1.0\textwidth]{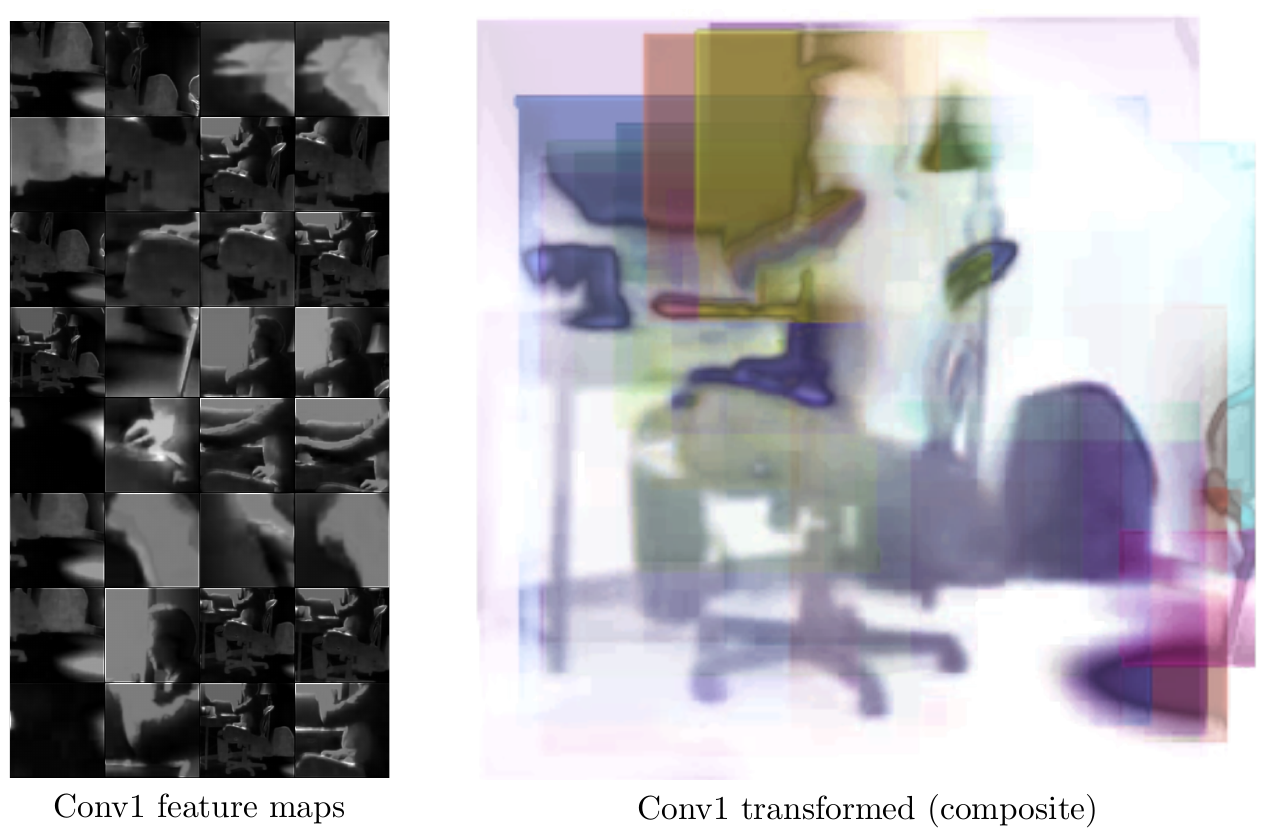}
\caption{Composite of 32 Conv1 feature maps from the fixation model. Different color indicates different frames.}
\label{fig:fixationplot}
\end{figure}

In Fig.~\ref{fig:fixationplot} we visualize frames from a Conv1 feature map applied to a video on a person sitting at their desk, since each feature has a transformation associated with it, even though some of the feature maps are at very small scales, the network can move them into a common coordinate frame as needed.

\section{Discussion \& Conclusion}
\label{sec:discussion}
We presented a framework that can use transformations to better use available data, moving beyond the implicit biases of the camera that recorded a video.

\paragraph{Video Stabilization.}
One of the hypotheses that we explore is stabilization in terms of a world frame, and in terms of a target (smooth pursuit). We discovered that stabilization on its own is not very helpful for current vision systems. However, when combined with fixation and saccades that highlight useful aspects of the input, it has significant advantages. 

\paragraph{Fixation in Animals.}
Similarly, eye movement and fixation in animals is hypothesized to be primarily in order to fully utilize the center of the retina, the fovea, that has high-resolution and only covers about 1-2 degrees of vision~\cite{provis2013adaptation}. This is, it is about efficient usage of resources. Furthermore, average fixation duration in humans is only 330 ms (8 frames at 24 fps)~\cite{henderson2003human}, suggesting long-term stabilization might be unnecessary for some visual systems. Thus eye movement primarily plays a role in allowing efficient visual search of the scene.

\paragraph{Stabilization in First-Person Videos.}
Stabilization in the first-person vision case is complicated by the fact that the camera is commonly doing smooth pursuit of the objects/hands of interest already, and zooming in is unnecessary because the field of view is already narrow. In the Charades third-person videos the camera often follows the person, that is, smooth pursuit is already present. 

\paragraph{Conclusion.}
We addressed several challenges that come with moving beyond the camera, and highlighted problems that have to be considered. Such as, variability due to scale and aspect ratio changes, how to include data augmentation, and working with pretrained networks that only consider particular types of input data.
We hope this work opens the door for systems that learn policies for visual search and efficient allocation of visual resources.

\vfill
{\small
\paragraph{Acknowledgements.}
This work was supported by Intel via the Intel Science and Technology Center for Visual Cloud Systems, Sloan Fellowship to AG, the Inria associate team GAYA, the ERC advanced grant ALLEGRO, gifts from Amazon and Intel, and the Indo-French project EVEREST (no.\ 5302-1) funded by CEFIPRA. The authors would like to thank Kenneth Marino for feedback on the manuscript.
}

\clearpage
\newpage
\section{Appendix}

\begin{figure}[b!]
    \centering
    \includegraphics[width=1.0\linewidth]{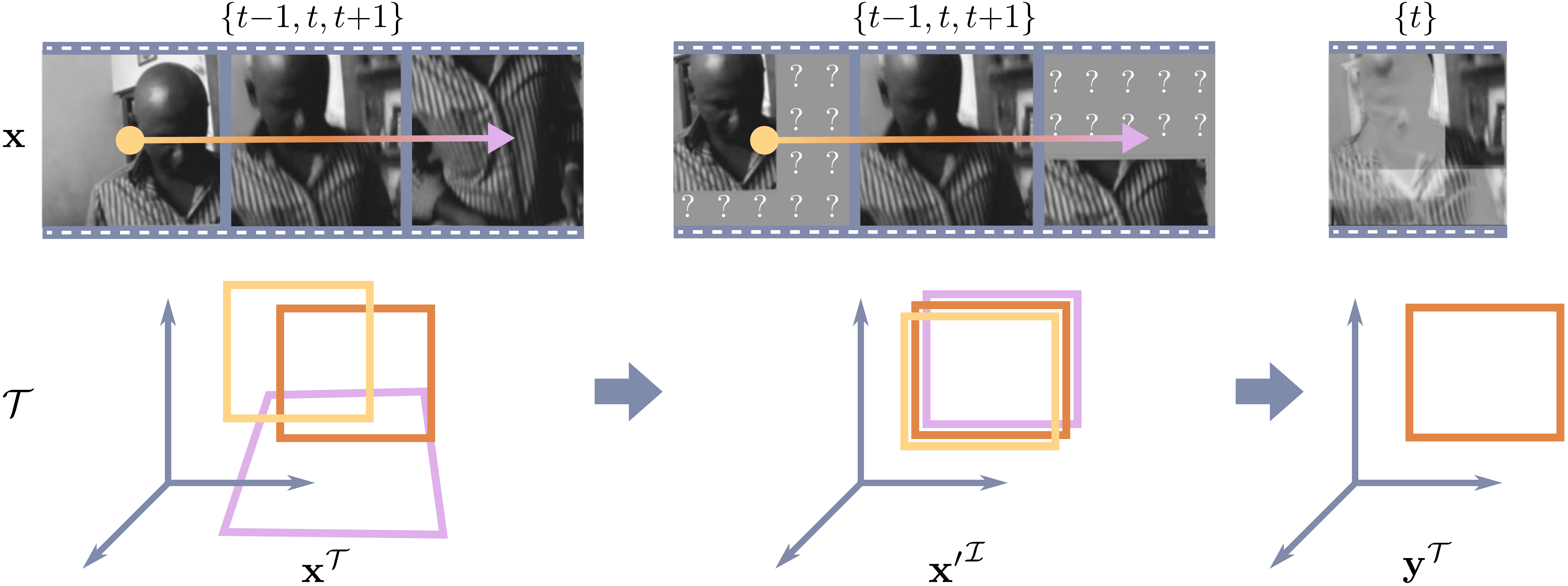}
    \caption{Illustration of our implementation of WorldFeatures. We start with a world feature map $\ve{x}^\mathcal{T}$ and and in order to calculate the output of a convolution applied to the feature map at time $t$, that is $\ve{y}^\mathcal{T}(t,h,w)$, we first transform the needed features into the coordinate system at time $t$, and then apply the convolution as normal. Black, gray and white, are used to denote $t{-}1$, $t$, and $t{+}1$, respectively.}
    \label{fig:worldconvolution}
\end{figure}

\paragraph{Implementation Details.}
Layers using WorldFeatures could be implemented with efficiency similar to their regular counterparts, requiring dedicated engineering. To iterate on the idea, we implemented WorldFeatures in high-level PyTorch code. We observe that current architectures have filter extent of 3 frames~\cite{he2016deep,wang2018non,i3d}. Thus, we simply transform each chunk of consecutive frames into common coordinates as needed for each filter computation. 
This is illustrated in Fig.~\ref{fig:worldconvolution}, where 3 frames are temporarily transformed into a common coordinate frame (the coordinates of frame $t$), and used to compute the convolution output.

Concretely, mirroring Eq.~\eqref{eq:sconv} above, suppose we need to calculate $\ve{y}^\mathcal{T}(t,h,w)$ for some particular frame: $t{=}t_0$, $h \in \left\{0{,}1{,}\dots{,}H{-}1\right\}$, and $w \in \left\{0{,}1{,}\dots{,}W{-}1\right\}$. Then we can rewrite as follows:
\begin{align}
\ve{x}'_t(t'{,}h'{,}w') &= \ve{x}(\mathcal{T}(t'{,}h'{,}w')) \\
\ve{y}^\mathcal{T}(t,h,w) &= \sum_i \sum_j \sum_k \ve{x}'_t(i{,}j{,}k) \cdot \ve{f}(t{-}i,h{-}j,w{-}k)
\end{align}
which describes convolution on $\ve{x}'$. Here $\ve{x}'$ is an explicitly transformed version of $\ve{x}$. Since $\ve{f}$ only has non-zero values at $t=\{{-}1,0,{+}1\}$ (if filter extent is 3) $\ve{x}'$ only needs to be computed for $t{-}1$, $t$, and $t{+}1$. Thus if we need to compute this for all values $T$ of $t$, we need to create $T$ such versions of $\ve{x}'$, which corresponds to 3 versions of the size of the input ($T {\times} H {\times} W$). Finally, we apply the regular convolution to these 3 copies and merge the results to create $\ve{y}^\mathcal{T}(t,h,w)$.

This has numerous advantages: 1) Since 3 or less frames are being transformed at a time, both cumulative errors in transformation and missing value problems are minimal. 2) Transforming across long timescales is only done in higher layers, where spatial resolution is lower, meaning less accuracy is needed. 3) Any layer that operates on the time dimension, such as maxpool, average pool, and 3d convolution, can be adapted in the same way using this method.

In Table~\ref{tab:arch} we illustrate 3DResNet50 extended with WorldFeatures, and how the size of the transformer changes to correspond to each layer. In our setup, all consecutive layers receive a (potentially resampled) transformer that contains the transformations to align consecutive timepoints of the feature map.
\newcommand{\blockb}[3]{\multirow{3}{*}{\(\left[\begin{array}{c}\text{1\x1, #2}\\[-.1em] \text{3\x3, #2}\\[-.1em] \text{1\x1, #1}\end{array}\right]\)\x#3}}
\begin{table}[tb]
\caption{Our 3DResNet50 model for video. The dimensions of 3D output maps and filter kernels are in T\x H\x W, with the number of channels following.
The input is 64\x224\x224, and the pre-computed transformer corresponding to the input is of size 64. Residual blocks are shown in brackets. The table is adapted from~\citet{wang2018non}.}
\label{tab:arch}
\vspace{-0.3cm}
\footnotesize
\centering
\resizebox{0.6\columnwidth}{!}{
\tablestyle{6pt}{1.08}
\begin{tabular}{c|c|c|c}
\multicolumn{2}{c|}{layer} & output size & $\begin{array}{c}\text{transformer}\\\text{output size}\end{array}$ \\
\shline
conv$_1$ & \multicolumn{1}{c|}{7\x7, 64, stride 2, 2, 2} & 32\x112\x112 & 32 \\
\hline
pool$_1$  & \multicolumn{1}{c|}{3\x3\x3 max, stride 2, 2, 2} & 16\x56\x56 & 16 \\
\hline
\multirow{3}{*}{res$_2$} & \blockb{256}{64}{3} & \multirow{3}{*}{16\x56\x56} & \multirow{3}{*}{16} \\
  &  & & \\
  &  & & \\
\hline
pool$_2$  & \multicolumn{1}{c|}{3\x1\x1 max, stride 2, 1, 1} & 8\x56\x56 & 8 \\
\hline
\multirow{3}{*}{res$_3$} & \blockb{512}{128}{4} & \multirow{3}{*}{8\x28\x28} & \multirow{3}{*}{8} \\
  &  & & \\
  &  & & \\
\hline
\multirow{3}{*}{res$_4$} & \blockb{1024}{256}{6} & \multirow{3}{*}{8\x14\x14} & \multirow{3}{*}{8}   \\
  &  &  & \\
  &  &  & \\
\hline
\multirow{3}{*}{res$_5$} & \blockb{2048}{512}{3} & \multirow{3}{*}{8\x7\x7} & \multirow{3}{*}{8}  \\
  &  & & \\
  &  & & \\
\hline
\multicolumn{2}{c|}{global average pool, fc} & 1\x1\x1 & 1 \\
\end{tabular}}
\vspace{-0.3cm}
\end{table}

\paragraph{Computing Stabilization.}
Before stabilization takes place, we need to compute the transformations between consecutive frames. Traditionally, optical flow or feature matching have been utilized for this purpose, but with increase in computation, recent work has moved towards direct methods for estimating the transformation. This optimizes the transformation directly on the image intensities, which enables utilizing all of the information in the image, making the estimation higher accuracy, and particularly robust to frames with few keypoints~\cite{engel2014lsd}.
In particular, we directly optimize the parameters of a transformation (affine or homography), but in a batch fashion on each pair of consecutive frames in each mini batch. The objective from LSD-SLAM~\cite{engel2014lsd} operates directly on images intensities (Huber norm) and downweighs large values. We simplify it to:
\begin{align}
    \min_\theta \sum_{h, w} \min \left( \left\lVert\ve{x}(t{,}h{,}w) - \ve{x}(\mathcal{T}_\theta(t{+}1{,}h{,}w))\right\rVert^2, \delta \right) \,,
\end{align}
where we align the frame at $t$ and $t{+}1$. We scale the source ($t+1$) and target ($t$) image equally such that target image has unit variance, and $\delta=0.01$. We then use gradient-based optimization with adaptive learning rate to minimize this objective with respect to the transformation parameters $\theta$.

\clearpage
\bibliographystyle{splncs04}
\bibliography{egbib}
\end{document}